\documentclass[11pt,a4paper]{article}
\usepackage[dvipsnames,table,xcdraw]{xcolor}
\usepackage[hyperref]{aacl-ijcnlp2020}
\usepackage{times}
\usepackage{latexsym}

\usepackage{array}
\newcolumntype{H}{>{\setbox0=\hbox\bgroup}c<{\egroup}@{}}

\usepackage{microtype}
\usepackage{graphicx}
\usepackage{times}
\usepackage{latexsym}
\usepackage{mathabx}
\usepackage{natbib}
\usepackage{tablefootnote}
\usepackage[mathletters]{ucs}
\usepackage[utf8x]{inputenc}
\usepackage{footmisc}
\usepackage{multirow}
\usepackage{multicol}
\usepackage{adjustbox}
\usepackage[english]{babel}

\aclfinalcopy 

\title{\textbf{Text-to-Text Pre-Training for Data-to-Text Tasks}}

\author{Mihir Kale \\
  Google Research\\
  \texttt{mihirkale@google.com} \\\And
  Abhinav Rastogi \\
  Google Research \\
  \texttt{abhirast@google.com} \\}

\date{}

\begin{document}
\maketitle
\begin{abstract}
We study the pre-train + fine-tune strategy for data-to-text tasks. Our experiments indicate that text-to-text pre-training in the form of T5 \cite{raffel2019exploring}, enables simple, end-to-end transformer based models to outperform pipelined neural architectures tailored for data-to-text generation, as well as alternative language model based pre-training techniques such as BERT and GPT-2. Importantly, T5 pre-training leads to better generalization, as evidenced by large improvements on out-of-domain test sets. We hope our work serves as a useful baseline for future research, as transfer learning becomes ever more prevalent for data-to-text tasks.
\end{abstract}

\section{Introduction}
Natural language generation from structured data, or data-to-text \cite{kukich1983design,mckeown1985text}, is the task of generating natural language text conditioned on source content provided in the form of structured data such as a table, graph etc. Some example applications include task oriented dialog \cite{wen2015semantically}, summarizing weather forecasts \cite{sripada2003sumtime, goldberg1994using},  etc.

In this work we study the applicability of large scale text-to-text transfer learning learning for this task. 
In particular, we focus on pre-training in the form of the “Text-to-Text Transfer Transformer” (T5) models released by \citet{raffel2019exploring}. Fine-tuning T5 achieves state-of-the-art results on  diverse benchmarks spanning 
task oriented dialogue (MultiWoz),
tables-to-text (ToTTo) and graph-to-text (WebNLG). Empirical results further demonstrate the following:
\begin{itemize}
    \item Pre-training greatly improves robustness of models to out-of-domain inputs.
    \item By leveraging pre-training, a simple end-to-end transformer model can outperform sophisticated, multi-stage pipelined approaches and other exotic architectures like graph neural networks.
    \item T5 outperforms alternatives like BERT \cite{devlin2018bert} and GPT-2 \cite{radford2019language}.
\end{itemize}
\par Our approach is simple, only scratching the surface of what is possible. There is much to be explored in the space of leveraging unlabelled data, developing unsupervised objectives etc. that are more tailored for generating text from structured data. We hope our work serves as a useful baseline for future research, as pre-training becomes ever more prevalent for this task. 

\section{Related Work}
\textbf{Data-to-Text} Early research on data-to-text focused on rule-based  methods \cite{reiter2000building}, while recent works have favored neural approaches \cite{wen2015semantically}.  
\citet{liu2018table} generate text by conditioning language models on tables,  \citet{puduppully2019data} explicitly model entities and \citet{marcheggiani2018deep} encode structured data using graph convolutional networks. \citet{ferreira2019neural} and \citet{moryossef2019step} find that neural pipelined approaches perform better than end-to-end models. \\
\textbf{Transfer Learning} \citet{devlin2018bert} showed that unsupervised pre-training can greatly benefit tasks like, question answering, summarization etc. In particular, \citet{raffel2019exploring} perform a large scale study of different training objectives, model capacity and size of data. \citet{peng2020few} and \citet{chen2019few} show that pre-training in the form of GPT-2 can indeed improve performance on the data-to-text task as well. 

\begin{figure}[h]
    \centering
    \includegraphics[width=\columnwidth,height=\textheight,keepaspectratio]{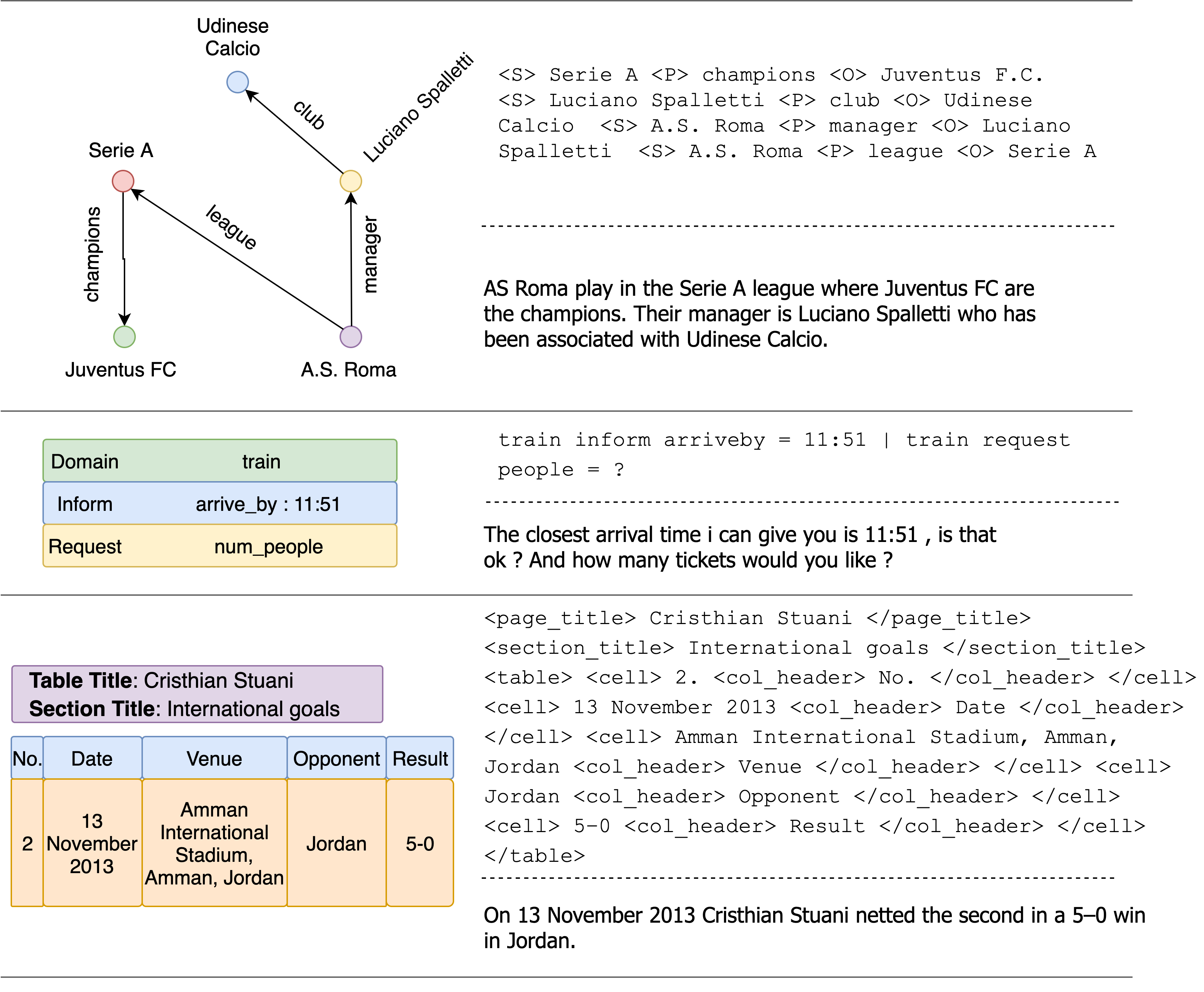}
    \caption{Examples from each dataset - The first row is WebNLG, second is Multiwoz and third is ToTTo. Each row illustrates the structured data (left), its linearized representation (top) and the target text(bottom)}
    \label{fig:d2t-examples}
\end{figure}

\section{Pre-training}
We rely on the T5 pre-trained models released by \citet{raffel2019exploring}. They consist of a transformer based encoder-decoder architecture. These models were pre-trained in a multitask fashion with an unsupervised “span masking” objective on Common Crawl data as well as supervised translation,
summarization, classification, and question answering tasks. Note that none of the supervised tasks include language generation from structured data. \footnote{Initial experiments with T5 variants trained on a purely unsupervised objective did not show any difference in performance.}
\par To study the impact of model capacity, we experiment  with different T5 variants - Small (60 million parameters), Base (220 million), Large (770 million) and 3B (3 billion).

\section{Fine-tuning}
Our modeling approach is simple. The data-to-text task is cast in the text-to-text framework by representing the structured data as a flat string (linearization). Figure \ref{fig:d2t-examples} shows examples of the input representation for each dataset. 
We then fine-tune T5 on the data-to-text corpus for a small number of steps. 
\par  Following  \cite{raffel2019exploring}, models are fine-tuned with a constant learning rate of 0.001. We use a batch size of 131,072 tokens, and a maximum input length of 512 tokens. The maximum training steps is set to 5K for  WebNLG, while the larger ToTTo dataset is trained for 10K steps. The T5 vocabulary consists of 32,000 sentencepieces. All the model parameters are updated in the fine-tuning process.

\par The best checkpoint is chosen based on the BLEU \cite{papineni2002bleu} score on the development set. Decoding is done via greedy search. In the final evaluation, for each dataset we rely on metrics used by prior work.

\section{Datasets}
We conduct experiments on 3 English datasets spanning a variety of domains.
\begin{itemize}
   \item \textbf{ToTTo} \cite{parikh2020totto} consists of Wikipedia tables paired with natural language descriptions. 
    The input is a set of cells from  a table, along with metadata such as the title of the table.
   \item \textbf{MultiWoz} \cite{budzianowski2018multiwoz} is a corpus of ~10K human-human dialogs for developing task oriented dialogue systems. For the NLG task, a meaning representation encapsulating system actions must be verbalized into natural language response. 
    \item \textbf{WebNLG} \cite{gardent2017webnlg}, where the task is to convert a graph of subject-object-predicate triples into a textual description.
\end{itemize}
\par Each dataset uses a different kind of structured data (tables, meaning representations and graph/triples).
Table \ref{dataset-stats} lists the sizes of the three datasets and Figure \ref{fig:d2t-examples} shows examples for each.

\begin{table}[h]
\centering
\begin{tabular}{l|l|l|l}  \hline
Dataset  & Train  & Dev  & Test \\ \hline
WebNLG   & 18.1K  & 2.2k & 4.9k \\
ToTTo    & 120K & 7.7k & 7.7k \\
Multiwoz & 56.8K  & 7.3k & 7.3k  \\ \hline
\end{tabular}
\caption{Dataset sizes.}
\label{dataset-stats}
\end{table}

\section{Results and Discussion}

\begin{table}[]
\centering
\begin{adjustbox}{width=\columnwidth,center}
\begin{tabular}{llll||lll} \hline
\multirow{2}{*}{Model} & \multicolumn{3}{c}{BLEU} & \multicolumn{3}{c}{METEOR}                            \\ 
                       & O  & S & U & O                  & S                     & U                   \\ \hline
 Melbourne$^\dagger$              & 45.1     & 54.5 & 33.3   & 0.37                     & 0.41                     & 0.33                     \\
 GTR-LSTM$^\dagger$              & 37.1     & 54.0 & 29.2   & 0.31                     & 0.37                     & 0.28                     \\ \hline
Pipe-Trans            & 51.7     & 56.4 & 38.9   & 0.32                     & 0.41                     & 0.21                     \\
Step$^\dagger$           & 47.4     & 53.3 & 34.4   & 0.39                     & 0.44                     & 0.34                     \\
DualEnc                & 51.4     & 63.4 & 36.7   & 0.41                     & 0.45                     & 0.37                     \\ \hline
 T5-Small               & 52.0     & 62.6 & 38.8   & 0.41 & 0.45 & 0.37 \\
 T5-Base                & 55.2     & \textbf{64.7} & 49.4   & 0.43 & \textbf{0.46} & 0.41 \\
T5-Large               & \textbf{57.1}     & 63.9 & \textbf{52.8}   & \textbf{0.44} & \textbf{0.46} & \textbf{0.41} \\ 
 T5-3B                  & 54.0     & 62.8 & 52.0   & 0.43 & 0.45 & 0.42 \\ \hline
\end{tabular}
\end{adjustbox}
\caption{Results on WebNLG. O stands for Overall test set, S for Seen and U for Unseen. Pipe-Trans is Pipeline-Transformer.}
\label{results-webnlg}
\end{table}

\subsection{WebNLG}
The evaluation is done using BLEU and METEOR \cite{lavie2007meteor}, similar to \cite{ferreira2019neural}. The test set is split into two parts - seen and unseen. The examples in the unseen set are drawn from domains not present in the training set, along with roughly 100 new predicates. \\
Some of the baselines we compare with are:
\begin{itemize}
    \item \textbf{Melbourne}, a neural encoder-decoder approach, which scored the highest in the automatic evaluation of the WebNLG challenge \cite{gardent2017webnlg}. The model relies on delexicalization, where entities are replaced with placeholders.
    \item \textbf{GTR-LSTM} \cite{distiawan2018gtr}, which employs a graph based triple encoder.
    \item \textbf{Step-by-Step} \cite{moryossef2019step} which splits the generation procedure into a planning stage followed by a neural generation stage.
    \item \textbf{Pipeline-Transformer} \cite{ferreira2019neural}, a pipelined neural system consisting of discourse ordering, text structuring, lexicalization and referring expression generation. 
    \item \textbf{DualEnc} \cite{zhao2020bridging}, the current state-of-the-art system. It consists of a graph convolution network based planning model which first predicts the order of the triples, followed by a separate LSTM with attention and copy mechanism model to generate the text. To train the planning model, the approach relies on extra annotations for the triple ordering. Such annotations are can be expensive and time consuming to obtain, especially for large, complex inputs. 
\end{itemize}
\par Results are reported in Table \ref{results-webnlg}, for the overall test set as well as the Seen and Unseen splits. T5-Large performs the best across BLEU as well as METEOR. It improves over DualEnc by 4.3 BLEU on the overall test set. It also displays excellent generalization to new domains and relations, with a 14 BLEU improvement on the unseen test set. The results indicate that with pre-training, end-to-end neural models can surpass sophisticated pipelined approaches while being much more robust to domain shift.

\begin{table}[h]
\centering
\begin{adjustbox}{width=\columnwidth,center}
\begin{tabular}{lccHHcc}
\hline
\multirow{2}{*}{Model} &  \multicolumn{2}{c}{Overall} & & & \multicolumn{2}{c}{Non-Overlap} \\
                  & BLEU        & PAR        & BLEU            & PAR           & BLEU             & PAR             \\ \hline
PGen & 41.6        & 51.6          & 50.6            & 58.0             & 32.2             & 45.2               \\
BERT-to-BERT      & 44.0        & 52.6          & 52.7            & 58.4             & 34.8             & 46.7               \\
T5-3B             & \textbf{49.5}        & \textbf{58.4}          & \textbf{57.5}            & \textbf{62.6}             & \textbf{41.4}             & \textbf{54.2}               \\ \hline
\end{tabular}
\end{adjustbox}
\caption{Results on the ToTTo test set. PAR is short for PARENT. PGen stands for Pointer Generetator \cite{see-etal-2017-get}.}
\label{results-totto}
\end{table}

\begin{table}[ht]
\centering
\begin{adjustbox}{width=\columnwidth,center}
\begin{tabular}{lllHHll} \hline
\multirow{2}{*}{Model} & \multicolumn{2}{c}{Overall} & &  & \multicolumn{2}{c}{Non-Overlap} \\ 
                      & BLEU        & PAR        & BLEU        & PAR        & BLEU          & PAR          \\ \hline
BERT-to-BERT      & 44.0        & 52.6          & 52.7            & 58.4             & 34.8             & 46.7   \\
T5-Small                  & 45.7        & 55.9         & 53.9        & 60.4         & 37.7          & 51.6           \\
T5-Base                   & 47.7        & 57.1         & 56.1        & 61.8         & 39.6          & 52.6           \\
T5-Large                  & 48.1        & 57.3         & 56.8        & 62.0         & 39.8          & 52.8           \\
T5-3B                     & 48.4        & 57.8          & 56.7        & 62.4          & 40.4          & 53.3      \\ \hline    
\end{tabular}
\end{adjustbox}
\caption{Results on the ToTTo development set for different variants of T5.}
\label{results-totto-dev}
\end{table}

\subsection{ToTTo}
Following \cite{parikh2020totto}, BLEU and PARENT are employed as evaluation metrics for this table-to-text generation task. PARENT is a reference less, word-overlap based metric that reflects the factual accuracy of generated text relative to the structured data. \citet{dhingra2019handling} find that PARENT correlates better with human factual accuracy judgements in comparison to other generation metrics like ROGUE \cite{lin2004rouge} and METEOR.  \\
The following baseline models are compared:
\begin{itemize}
    \item \textbf{Pointer Generator} \cite{see2017get} - An LSTM based seq2seq model with attention and pointer network based copy mechanism.
    \item \textbf{BERT-to-BERT} \cite{rothe2019leveraging} - A transformer based encoder-decoder model, where both the encoder and decoder are initialized with BERT.
\end{itemize}
\par Since it deals with open domain tables, ToTTo is arguably the most challenging dataset. Notably, it features a hidden test set, which is split into two halves - Overlap and Non-Overlap. The Non-Overlap test set features examples that are out-of-domain from the training set. 
\par Results are reported in Table \ref{results-totto}. T5-3B\footnote{We used beam search with a width of 10 for the test set submission.} achieves state-of-the-art results \footnote{The leaderboard can be found at https://github.com/google-research-datasets/totto.}, improving upon the BERT baseline by 5.5 BLEU and 5.8 PARENT. Moreover, the model is more robust to out-of-domain tables, with larger improvements of 6.6 BLEU and 7.5 PARENT on the Non-Overlap test set. Table \ref{results-totto-dev} reports results on the development set for the different T5 model sizes. 
T5-Small outperforms BERT-to-BERT, even though it has 3x fewer parameters (220M vs 60M).

\begin{table}[h]
\centering
\begin{tabular}{lll} \hline
Model    & BLEU                         & SER                         \\ \hline
HDSA$^\dagger$      & 26.5                        & 12.14                       \\
SC-GPT2  & 30.8                        & \textbf{0.53}                        \\ \hline
T5-Small & 34.6  & 1.27 \\
T5-Base  & \textbf{35.1} & 0.99 \\ 
T5-Large & 34.7 & 0.92 \\
T5-3B    & 34.8 & 0.86 \\ \hline
\end{tabular}
\caption{Results on Multiwoz. $^\dagger$\cite{chen2019semantically} }
\label{results-multiwoz}
\end{table}

\subsection{MultiWoz}
Evaluation on MultiWoz is done using BLEU and SER (Slot Error Rate). SER is the fraction of examples where at least one slot value from the structured data is not expressed in the predicted response. \footnote{The metric is noisy since the comparison is done via exact match, does not accoutn for paraphrases and does not cover all slots.} 
\par Our baselines are 
\begin{itemize}
    \item \textbf{HDSA} \cite{chen2019semantically} is a transformer based architecture that encodes the dialog acts into a multi-layer hierarchical graph, with individual attention heads modeling specific nodes in graph. 
    \item \textbf{SC-GPT2}  \cite{peng2020few} is a GPT-2 (345M parameters) model that is further pre-trained on a large data-to-text dialog corpus consisting of 400,000 examples and finally fine-tuned on MultiWoz. This 2 stage pre-training approach is currently state-of-the-art for Multiwoz.
\end{itemize} 

\par Results are reported in Table \ref{results-multiwoz}. All T5 based models (including T5-small which has 5x fewer parameters) outperform SC-GPT2 by 4-5 BLEU without any in-domain pre-training. We note that the SER score on MultiWOZ is slightly worse in comparison with SC-GPT.  SC-GPT generates 5 predictions for each input and then ranks them based on the SER score itself, which naturally leads to better slot error rates. On the other hand, we generate a single output.

\begin{table}[ht]
\centering
\begin{tabular}{l|ll|ll}
\hline
& \multicolumn{2}{c|}{Seen} & \multicolumn{2}{c}{Unseen}  \\
& Nat & Acc & Nat & Acc \\\hline
DualEnc & 2.30 & 89.2 & 1.99 & 66 \\
T5-Large &  \textbf{2.39} & \textbf{92.0} & \textbf{2.33} & \textbf{90.0} \\ \hline 
\end{tabular}
\caption{Human evaluation on WebNLG. Nat is short for Naturalness and Acc is short for Accuracy.}
\label{results-human-webnlg}
\end{table}

\begin{table*}[h]
\begin{adjustbox}{width=\textwidth,center}
\begin{tabular}{|l|l|} \hline
Input & \textless{}aidastella, christening date, 2013-03-16\textgreater{}                                                                                                  \\
DualEnc & Aidastella was \textcolor{red}{inaugurated} on March 16 , 2013 .                                                                                                                    \\  
T5      & Aidastella was \textcolor{ForestGreen}{christened} on March 16 , 2013 .                                                                                                                     \\ \hline
Input  & \textless{}Andra (singer).  genre , rhythm and blues\textgreater{}                                                                                                 \\
DualEnc & Andra singer is rhythm and blues .                                                                                                                                 \\
T5      & Andra is a singer who plays rhythm and blues .                                                                                                                     \\ \hline
Input & \textless{}Aaron deer, genre, indie rock\textgreater \textless{}Aaron Deer, origin, Indiana\textgreater \textless{}Aaron Deer, origin, United States\textgreater{} \\
DualEnc & Aaron Deer , indie rock , has a origin of Indiana and is located in United States .                                                                                \\
T5      & Aaron Deer is an American from Indiana who is part of the genre of indie rock .            \\ \hline   
Input & \textless{}Alvah Sabin, birth date, 1793-10-23\textgreater \textless{}Alvah Sabin, office (worked at , worked as), secretary of state of Vermont\textgreater{}	\\
DualEnc & Alvah Sabin was born on October 23 , 1793 and \textcolor{red}{is in} secretary of state of Vermont . \\
T5 & Alvah Sabin was born on 23 October 1793 and \textcolor{ForestGreen}{served as} secretary of state of Vermont .	 \\ \hline   
\end{tabular}
\end{adjustbox}
\caption{Model predictions on the WebNLG Unseen set. DualEnc struggles to verbalize predicates and produces ungrammatical output. T5 output is accurate and more grammatical.}
\label{model-predictions}
\end{table*}

\subsection{Human Evaluation}
We conduct a human evaluation study on WebNLG. Human raters are presented with predicted text, along with up to 3 ground truth references. They are asked to judge the prediction along two axes - (1) Accuracy - A binary rating to gauge whether the prediction conveys the same information as the gold references and (2) Naturalness - A five point scale between 1-3, with 3 indicating a perfectly fluent and grammatical response. Each prediction is rated by 3 raters. For accuracy, we take the majority vote and for naturalness we take the average.
We evaluate 500 examples, equally split between the Seen and Unseen test sets.
\par The evaluation is performed for T5-Large and the current state-of-the-art DualEnc model. Results are reported in Table \ref{results-human-webnlg}. On the Seen set, both models perform well, with T5 being rated  better across both metrics. On the Unseen set, DualEnc shows a large drop of 24\% in accuracy while the fluency degrades to just 1.99. Remarkably, T5 sees only a marginal drop, scoring 90\% on accuracy and 2.33 on fluency. Table \ref{model-predictions} shows some qualitative examples.

\subsection{Impact of model capacity}
Our experiments with different T5 variants of varying sizes shed some light on how model capacity impacts performance. The results suggest that it largely depends on the size and complexity of the dataset. 
For instance, MultiWoz exhibits the least variation in the structured data and is fairly large at 56k examples. Here, even the smallest model T5-Small, is on par with the larger models. 
WebNLG has only 18K examples and features roughly 200 distinct relations. On the seen test set, all models perform comparably. However, on the unseen test set we notice that performance increases with model size. In particular, there is a stark jump of 10 BLEU when going from T5-Small to T5-Base, implying that model capacity is critical for out-of-domain generalization. A similar trend is observed for ToTTo (Table \ref{results-totto-dev}), with a noticeable improvement from Small to Base, followed by smaller improvements upto T5-3B.

\section{Conclusion}
In this study we evaluated pre-training in the form of T5 for the data-to-text task. We found that it leads to state-of-the-art results, while greatly improving robustness to out-of-domain inputs. In the future, we hope to design unsupervised pre-training objectives that are specifically tailored for the data-to-text task. 
We also hope to extend this work to multiple languages, especially low resource ones.

\bibliography{anthology,aacl-ijcnlp2020}

\begin{thebibliography}{29}
\expandafter\ifx\csname natexlab\endcsname\relax\def\natexlab#1{#1}\fi

\bibitem[{Budzianowski et~al.(2018)Budzianowski, Wen, Tseng, Casanueva, Ultes,
  Ramadan, and Gasic}]{budzianowski2018multiwoz}
Pawe{\l} Budzianowski, Tsung-Hsien Wen, Bo-Hsiang Tseng, I{\~n}igo Casanueva,
  Stefan Ultes, Osman Ramadan, and Milica Gasic. 2018.
\newblock Multiwoz-a large-scale multi-domain wizard-of-oz dataset for
  task-oriented dialogue modelling.
\newblock In \emph{Proceedings of the 2018 Conference on Empirical Methods in
  Natural Language Processing}, pages 5016--5026.

\bibitem[{Chen et~al.(2019{\natexlab{a}})Chen, Chen, Qin, Yan, and
  Wang}]{chen2019semantically}
Wenhu Chen, Jianshu Chen, Pengda Qin, Xifeng Yan, and William~Yang Wang.
  2019{\natexlab{a}}.
\newblock Semantically conditioned dialog response generation via hierarchical
  disentangled self-attention.
\newblock In \emph{Proceedings of the 57th Annual Meeting of the Association
  for Computational Linguistics}, pages 3696--3709.

\bibitem[{Chen et~al.(2019{\natexlab{b}})Chen, Eavani, Liu, and
  Wang}]{chen2019few}
Zhiyu Chen, Harini Eavani, Yinyin Liu, and William~Yang Wang.
  2019{\natexlab{b}}.
\newblock Few-shot nlg with pre-trained language model.
\newblock \emph{arXiv preprint arXiv:1904.09521}.

\bibitem[{Devlin et~al.(2018)Devlin, Chang, Lee, and
  Toutanova}]{devlin2018bert}
Jacob Devlin, Ming-Wei Chang, Kenton Lee, and Kristina Toutanova. 2018.
\newblock Bert: Pre-training of deep bidirectional transformers for language
  understanding.
\newblock \emph{arXiv preprint arXiv:1810.04805}.

\bibitem[{Dhingra et~al.(2019)Dhingra, Faruqui, Parikh, Chang, Das, and
  Cohen}]{dhingra2019handling}
Bhuwan Dhingra, Manaal Faruqui, Ankur Parikh, Ming-Wei Chang, Dipanjan Das, and
  William Cohen. 2019.
\newblock Handling divergent reference texts when evaluating table-to-text
  generation.
\newblock In \emph{Proceedings of the 57th Annual Meeting of the Association
  for Computational Linguistics}, pages 4884--4895.

\bibitem[{Distiawan et~al.(2018)Distiawan, Qi, Zhang, and
  Wang}]{distiawan2018gtr}
Bayu Distiawan, Jianzhong Qi, Rui Zhang, and Wei Wang. 2018.
\newblock Gtr-lstm: A triple encoder for sentence generation from rdf data.
\newblock In \emph{Proceedings of the 56th Annual Meeting of the Association
  for Computational Linguistics (Volume 1: Long Papers)}, pages 1627--1637.

\bibitem[{Ferreira et~al.(2019)Ferreira, van~der Lee, van Miltenburg, and
  Krahmer}]{ferreira2019neural}
Thiago~Castro Ferreira, Chris van~der Lee, Emiel van Miltenburg, and Emiel
  Krahmer. 2019.
\newblock Neural data-to-text generation: A comparison between pipeline and
  end-to-end architectures.
\newblock In \emph{Proceedings of the 2019 Conference on Empirical Methods in
  Natural Language Processing and the 9th International Joint Conference on
  Natural Language Processing (EMNLP-IJCNLP)}, pages 552--562.

\bibitem[{Gardent et~al.(2017)Gardent, Shimorina, Narayan, and
  Perez-Beltrachini}]{gardent2017webnlg}
Claire Gardent, Anastasia Shimorina, Shashi Narayan, and Laura
  Perez-Beltrachini. 2017.
\newblock The webnlg challenge: Generating text from rdf data.
\newblock In \emph{Proceedings of the 10th International Conference on Natural
  Language Generation}, pages 124--133.

\bibitem[{Goldberg et~al.(1994)Goldberg, Driedger, and
  Kittredge}]{goldberg1994using}
Eli Goldberg, Norbert Driedger, and Richard~I Kittredge. 1994.
\newblock Using natural-language processing to produce weather forecasts.
\newblock \emph{IEEE Expert}, 9(2):45--53.

\bibitem[{Kukich(1983)}]{kukich1983design}
Karen Kukich. 1983.
\newblock Design of a knowledge-based report generator.
\newblock In \emph{Proceedings of the 21st annual meeting on Association for
  Computational Linguistics}, pages 145--150. Association for Computational
  Linguistics.

\bibitem[{Lavie and Agarwal(2007)}]{lavie2007meteor}
Alon Lavie and Abhaya Agarwal. 2007.
\newblock Meteor: An automatic metric for mt evaluation with high levels of
  correlation with human judgments.
\newblock In \emph{Proceedings of the Second Workshop on Statistical Machine
  Translation}, pages 228--231. Association for Computational Linguistics.

\bibitem[{Lin(2004)}]{lin2004rouge}
Chin-Yew Lin. 2004.
\newblock Rouge: A package for automatic evaluation of summaries.
\newblock In \emph{Text summarization branches out}, pages 74--81.

\bibitem[{Liu et~al.(2018)Liu, Wang, Sha, Chang, and Sui}]{liu2018table}
Tianyu Liu, Kexiang Wang, Lei Sha, Baobao Chang, and Zhifang Sui. 2018.
\newblock Table-to-text generation by structure-aware seq2seq learning.
\newblock In \emph{Thirty-Second AAAI Conference on Artificial Intelligence}.

\bibitem[{Marcheggiani and Perez-Beltrachini(2018)}]{marcheggiani2018deep}
Diego Marcheggiani and Laura Perez-Beltrachini. 2018.
\newblock Deep graph convolutional encoders for structured data to text
  generation.
\newblock In \emph{Proceedings of the 11th International Conference on Natural
  Language Generation}, pages 1--9.

\bibitem[{McKeown(1985)}]{mckeown1985text}
Kathleen~R McKeown. 1985.
\newblock Text generation: using discourse strategies and focus constraints to
  generate natural language text.

\bibitem[{Moryossef et~al.(2019)Moryossef, Goldberg, and
  Dagan}]{moryossef2019step}
Amit Moryossef, Yoav Goldberg, and Ido Dagan. 2019.
\newblock Step-by-step: Separating planning from realization in neural
  data-to-text generation.
\newblock In \emph{Proceedings of the 2019 Conference of the North American
  Chapter of the Association for Computational Linguistics: Human Language
  Technologies, Volume 1 (Long and Short Papers)}, pages 2267--2277.

\bibitem[{Papineni et~al.(2002)Papineni, Roukos, Ward, and
  Zhu}]{papineni2002bleu}
Kishore Papineni, Salim Roukos, Todd Ward, and Wei-Jing Zhu. 2002.
\newblock Bleu: a method for automatic evaluation of machine translation.
\newblock In \emph{Proceedings of the 40th annual meeting on association for
  computational linguistics}, pages 311--318. Association for Computational
  Linguistics.

\bibitem[{Parikh et~al.(2020)Parikh, Wang, Gehrmann, Faruqui, Dhingra, Yang,
  and Das}]{parikh2020totto}
Ankur~P Parikh, Xuezhi Wang, Sebastian Gehrmann, Manaal Faruqui, Bhuwan
  Dhingra, Diyi Yang, and Dipanjan Das. 2020.
\newblock Totto: A controlled table-to-text generation dataset.
\newblock \emph{arXiv preprint arXiv:2004.14373}.

\bibitem[{Peng et~al.(2020)Peng, Zhu, Li, Li, Li, Zeng, and Gao}]{peng2020few}
Baolin Peng, Chenguang Zhu, Chunyuan Li, Xiujun Li, Jinchao Li, Michael Zeng,
  and Jianfeng Gao. 2020.
\newblock Few-shot natural language generation for task-oriented dialog.
\newblock \emph{arXiv preprint arXiv:2002.12328}.

\bibitem[{Puduppully et~al.(2019)Puduppully, Dong, and
  Lapata}]{puduppully2019data}
Ratish Puduppully, Li~Dong, and Mirella Lapata. 2019.
\newblock Data-to-text generation with content selection and planning.
\newblock In \emph{Proceedings of the AAAI Conference on Artificial
  Intelligence}, volume~33, pages 6908--6915.

\bibitem[{Radford et~al.(2019)Radford, Wu, Child, Luan, Amodei, and
  Sutskever}]{radford2019language}
Alec Radford, Jeffrey Wu, Rewon Child, David Luan, Dario Amodei, and Ilya
  Sutskever. 2019.
\newblock Language models are unsupervised multitask learners.
\newblock \emph{OpenAI Blog}, 1(8).

\bibitem[{Raffel et~al.(2019)Raffel, Shazeer, Roberts, Lee, Narang, Matena,
  Zhou, Li, and Liu}]{raffel2019exploring}
Colin Raffel, Noam Shazeer, Adam Roberts, Katherine Lee, Sharan Narang, Michael
  Matena, Yanqi Zhou, Wei Li, and Peter~J Liu. 2019.
\newblock Exploring the limits of transfer learning with a unified text-to-text
  transformer.
\newblock \emph{arXiv preprint arXiv:1910.10683}.

\bibitem[{Reiter and Dale(2000)}]{reiter2000building}
Ehud Reiter and Robert Dale. 2000.
\newblock \emph{Building natural language generation systems}.
\newblock Cambridge university press.

\bibitem[{Rothe et~al.(2019)Rothe, Narayan, and Severyn}]{rothe2019leveraging}
Sascha Rothe, Shashi Narayan, and Aliaksei Severyn. 2019.
\newblock Leveraging pre-trained checkpoints for sequence generation tasks.
\newblock \emph{arXiv preprint arXiv:1907.12461}.

\bibitem[{See et~al.(2017{\natexlab{a}})See, Liu, and
  Manning}]{see-etal-2017-get}
Abigail See, Peter~J. Liu, and Christopher~D. Manning. 2017{\natexlab{a}}.
\newblock \href {https://doi.org/10.18653/v1/P17-1099} {Get to the point:
  Summarization with pointer-generator networks}.
\newblock In \emph{Proceedings of the 55th Annual Meeting of the Association
  for Computational Linguistics (Volume 1: Long Papers)}, pages 1073--1083,
  Vancouver, Canada. Association for Computational Linguistics.

\bibitem[{See et~al.(2017{\natexlab{b}})See, Liu, and Manning}]{see2017get}
Abigail See, Peter~J Liu, and Christopher~D Manning. 2017{\natexlab{b}}.
\newblock Get to the point: Summarization with pointer-generator networks.
\newblock In \emph{Proceedings of the 55th Annual Meeting of the Association
  for Computational Linguistics (Volume 1: Long Papers)}, pages 1073--1083.

\bibitem[{Sripada et~al.()Sripada, Reiter, and Davy}]{sripada2003sumtime}
Somayajulu Sripada, Ehud Reiter, and Ian Davy.
\newblock Sumtime-mousam: Configurable marine weather forecast generator.

\bibitem[{Wen et~al.(2015)Wen, Gasic, Mrksic, Su, Vandyke, and
  Young}]{wen2015semantically}
Tsung-Hsien Wen, Milica Gasic, Nikola Mrksic, Pei-Hao Su, David Vandyke, and
  Steve Young. 2015.
\newblock Semantically conditioned lstm-based natural language generation for
  spoken dialogue systems.
\newblock \emph{arXiv preprint arXiv:1508.01745}.

\bibitem[{Zhao et~al.(2020)Zhao, Walker, and Chaturvedi}]{zhao2020bridging}
Chao Zhao, Marilyn Walker, and Snigdha Chaturvedi. 2020.
\newblock Bridging the structural gap between encoding and decoding for
  data-to-text generation.
\newblock In \emph{Proceedings of the 58th Annual Meeting of the Association
  for Computational Linguistics (Volume 1: Long Papers)}.

\end{thebibliography}
\bibliographystyle{acl_natbib}


\end{document}